\pdfoutput=1

\documentclass[11pt]{article}

\usepackage[rgb,dvipsnames]{xcolor}
\usepackage{acl}

\usepackage{times}
\usepackage{latexsym}
\usepackage{booktabs}
\usepackage{multirow}
\usepackage{amsmath, amsthm, amssymb}
\usepackage{cleveref}
\usepackage{longtable}
\usepackage{tikz}

\usepackage[T1]{fontenc}

\usepackage[utf8]{inputenc}

\usepackage{microtype}
\usepackage{hyperref}
\usepackage{graphics}
\usepackage{mwe}

\usepackage{arydshln}

\newif\ifprintcomments

\DeclareMathOperator*{\argmax}{arg\,max}

\newcommand{\x}{{\bf x}}

\newcommand{\btheta}{\boldsymbol{\theta}}
\newcommand{\bphi}{\boldsymbol{\phi}}

\newcommand{\modelname}{{\sc ATInteR}\xspace}
\title{Don’t Retrain, Just Rewrite: Countering Adversarial Perturbations \\ by Rewriting Text}

\author{
Ashim Gupta\textsuperscript{\rm 1}\thanks{~ Work done during internship at Bloomberg} ,
Carter Wood Blum\textsuperscript{\rm 2},
Temma Choji\textsuperscript{\rm 2},\\
\bf 
Yingjie Fei\textsuperscript{\rm 2},
Shalin Shah\textsuperscript{\rm 2},
Alakananda Vempala\textsuperscript{\rm 2},\\
\bf
Vivek Srikumar\textsuperscript{\rm 1}\\ 
\textsuperscript{\rm 1}University of Utah,
\textsuperscript{\rm 2}Bloomberg,\\
\{ashim, svivek\}@cs.utah.edu, \\
\{szhang611, cblum18, yfei29, sshah804, tchoji, avempala\}@bloomberg.net \\
}

\begin{document}
\printcommentstrue 
\maketitle
\begin{abstract}
Can language models transform inputs to protect text classifiers against adversarial attacks? 
In this work, we present~\modelname, a model that intercepts and
learns to rewrite adversarial inputs to make them non-adversarial for a downstream text classifier. 
Our experiments on four datasets and five attack mechanisms reveal that \modelname\ is  effective at providing better adversarial robustness than existing defense approaches, without compromising task accuracy.
For example, on sentiment classification using the SST-2 dataset, our method improves the adversarial accuracy over the best existing defense approach by more than 4\% with a smaller decrease in task accuracy (0.5 \% vs. 2.5\%). 
Moreover, we show that \modelname\ generalizes across multiple downstream tasks and classifiers without having to explicitly retrain it for those settings. Specifically, we find that when~\modelname\ is trained to remove adversarial perturbations for the sentiment classification task on the SST-2 dataset, it even transfers to a semantically different task of news classification (on AGNews) and 
improves the adversarial robustness by more than 10\%.
\end{abstract}

\section{Introduction}
\label{sec:intro}
\begin{figure*}
\includegraphics[width=\textwidth]{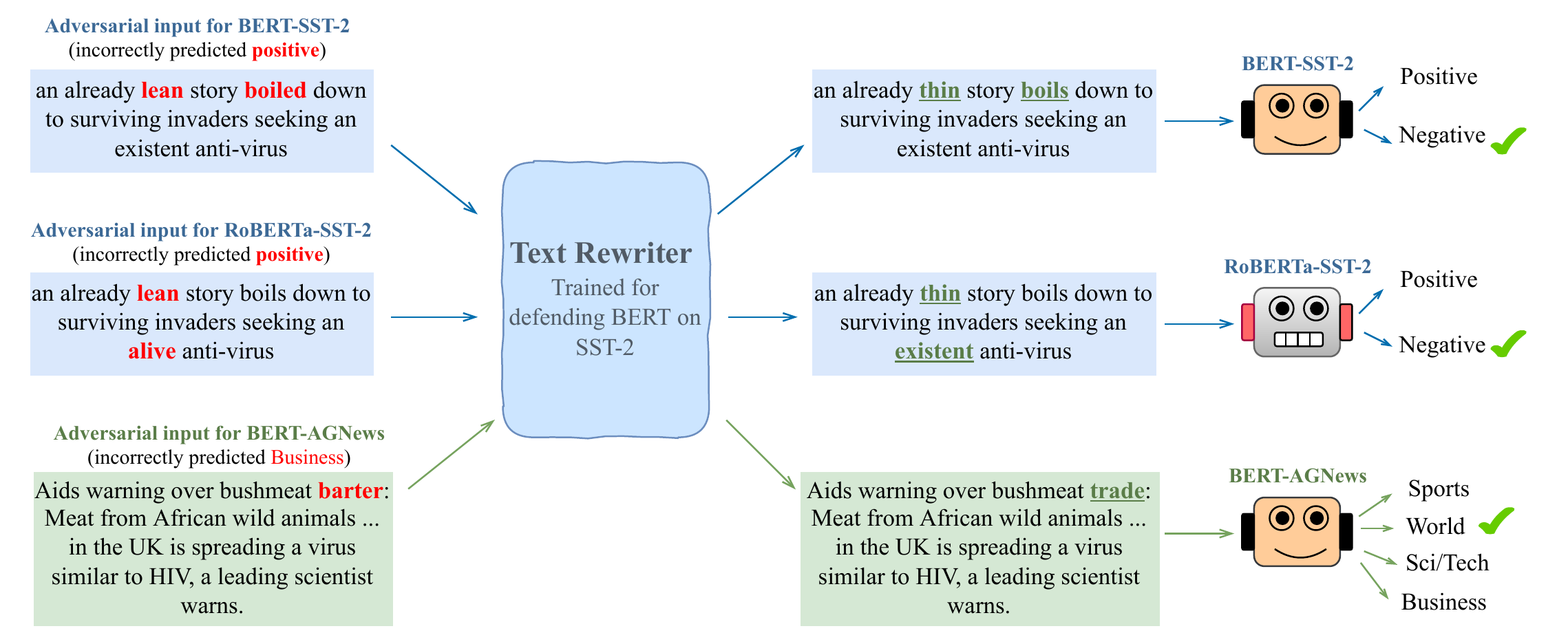}
\caption{Modular application of ~\modelname, which is trained for defending a BERT classifier for the SST-2 dataset. Demonstrating transferability across models, ~\modelname\ successfully defends a RoBERTa classifier on SST-2 without retraining. Similarly,~\modelname\ is successful in defending a BERT model for a news classification task on AGNews. Refer to~\cref{sec:results} for a more detailed discussion. 
}
\label{fig:modular}
\end{figure*}
Neural models in NLP have been shown to be vulnerable to adversarial attacks both during 
training time~\citep{gu2017badnets,wallace-etal-2019-universal,wallace-etal-2021-concealed,https://doi.org/10.48550/arxiv.2110.08247} and at deployment~\citep{ebrahimi-etal-2018-hotflip,jin2020bert,garg-ramakrishnan-2020-bae}. 
The attacks of the latter type aim to craft adversarial inputs by introducing small, imperceptible perturbations in the input text that erroneously change the output label of a classification model. 
Defending against such attacks is important because it ensures the integrity and reliability of NLP systems. 
If undefended, for example, an attacker could adversarially manipulate a spam email to evade a spam detector.

An ideal defense mechanism against such adversarial attacks should maintain good task performance on non-adversarial inputs, effectively mitigate adversarial attacks, and be transferable to other models and datasets. 
The transferability of defenses is a valuable property because it allows easy application to new and unknown models without retraining the underlying classification model.
This is particularly useful when complete access to the model is not possible; for example when the model is accessed through an API.
Most existing methods do not satisfy these desiderata, typically lacking in one or more desired properties.
For example, the methods that use input randomization like SAFER~\citep{ye-etal-2020-safer} and Sample Shielding~\citep{rusert-srinivasan-2022-dont} significantly degrade task accuracies due to the smoothing and aggregation involved, and are thus ineffective  defenses. 
Another set of methods--- e.g., adversarial training~\citep{jia-liang-2017-adversarial,ebrahimi-etal-2018-hotflip} and SHIELD~\citep{le-etal-2022-shield}---require model retraining; while serving as effective defenses, they cannot be transferred to other models and datasets without retraining the classifier.

In this work, we present a novel strategy for defending against adversarial attacks that satisfies the aforementioned desiderata. Our method---\textbf{A}dversarial \textbf{T}ext \textbf{Inte}rceptor and \textbf{R}ewriter\ (~\modelname)---is based on 
the intuition that automatically generated adversarial inputs can be undone by \emph{learning} to manipulate
the textual inputs instead of retraining the classification model. 
Specifically, we employ an encoder-decoder module that intercepts and rewrites the adversarial input to remove any adversarial perturbations before feeding it to the classifier\footnote{By \textit{intercepting}, we simply mean that ~\modelname\ stops the input from directly going to the classifier and processes it first to remove the adversarial perturbations. 
}. 
%
%
%
Our method differs from existing input randomization approaches in that it does not rely on random word replacements or deletions to counteract adversarial changes. Instead, we employ a separate model that is explicitly trained to remove adversarial perturbations. One benefit of this strategy is that it dissociates the responsibility of ensuring adversarial robustness from the classification model and delegates it to an external module, the text rewriter. Consequently, the rewriter module serves as a pluggable component, enabling it to defend models that it was not explicitly trained to protect.~\Cref{fig:modular} demonstrates this scenario.  

We demonstrate the effectiveness of our approach using a T5 model~\citep{raffel2020exploring} as the general-purpose text rewriter, but our method is applicable to any transformer-based text generator/rewriter. 
Through extensive experimentation and comparison with existing methods, we show that~\modelname\ effectively removes adversarial perturbations, and consistently outperforms other defense approaches on several datasets for text classification. 
When used as a pluggable component,~\modelname\ exhibits good transferability to new models and datasets without 
the need for retraining
(examples shown in~\cref{fig:modular}).
Specifically, we find that this T5-based rewriter trained to remove adversarial perturbations for the sentiment classification task on the SST-2 dataset, also 
removes adversarial perturbations for a news classification model (on AGNews),  increasing adversarial robustness by over 10\%. 

In summary, our contributions are:
\begin{enumerate}
    \item We propose a novel defense mechanism against adversarial attacks, called ~\modelname\, that uses a text rewriter module, along with a simple strategy to train this module.

    \item We demonstrate its effectiveness on four benchmark datasets and five adversarial attacks. Compared with competitive baselines, our method substantially improves the adversarial robustness with a much smaller decrease in accuracy on non-adversarial inputs. 

    \item We show that ~\modelname\ can be used as a pluggable module without retraining and that its ability to defend models is transferable to new models (e.g., BERT $\rightarrow$ RoBERTa) as well as new datasets (BERT on SST-2 $\rightarrow$ BERT on AGNews). 
    
\end{enumerate}

\section{Related Work}
\label{sec:related_work}
\paragraph{Adversarial Attacks } 
Most adversarial attacks use heuristic-based search methods to substitute vulnerable parts of the input with carefully chosen adversarial text~\citep{ebrahimi-etal-2018-hotflip,jin2020bert,jia-liang-2017-adversarial}. These substitutions can be performed at the character-level~\citep{gao2018black,belinkov2018synthetic}, word-level~\citep{ren-etal-2019-generating,jin2020bert,garg2020bae,li-etal-2020-bert-attack,zhang-etal-2021-crafting}, or both~\citep{li2018textbugger}. Finally, while adversarial attacks show that NLP models are over-sensitive to small perturbations, NLP models have also been shown to be under-sensitive to certain perturbations like input reduction, etc.~\citep{feng-etal-2018-pathologies,gupta2021bert}. 
We refer the reader to the detailed recent survey of~\citet{wang-etal-2022-measure}. 

\paragraph{Defenses against Adversarial Attacks } The typical strategies employed for defending text classification systems against adversarial attacks involve either retraining the classifiers using adversarial examples or incorporating randomized smoothing at the input stage to make robust predictions. The defenses of the former type involve \textit{adversarial training}~\citep{goodfellow2015explaining,alzantot-etal-2018-generating}, \textit{certified training}~\citep{jia-etal-2019-certified,zhou-etal-2021-defense,huang-etal-2019-achieving}, and other specialized training schemes~\citep{le-etal-2022-shield,jiang2022rose}. While adversarial training lacks in effectiveness~\citep{alzantot-etal-2018-generating}, the certification based methods are only applicable for a specific set of perturbations; e.g., ~\citet{jia-etal-2019-certified} restrict word substitutions to belong to a counter-fitted embedding space~\citep{mrksic-etal-2016-counter}. More recently, ~\citet{le-etal-2022-shield} proposed SHIELD, that trains a stochastic ensemble of experts by only \textit{patching} the last layer of the BERT classifier. We use SHIELD, and adversarial training for comparison with our proposed method.

On the other end of the spectrum are the models that do not retrain the classifier and instead use randomized smoothing techniques to enhance the robustness of the models~\citep{cohen2019certified,zhou-etal-2019-learning,ye-etal-2020-safer,rusert-srinivasan-2022-dont,wang-etal-2022-distinguishing}.~\citet{ye-etal-2020-safer} introduced a defense called SAFER, that significantly improves certified robustness by performing randomized substitutions using a synonym network.~\citet{rusert-srinivasan-2022-dont} proposed another randomization defense called Sample Shielding, which relies on making an ensemble of predictions on different random samples of the input text. 
One major drawback of utilizing randomization-based techniques is that they may result in a significant decrease in task accuracies on non-adversarial inputs. To overcome this limitation, ~\citet{bao-etal-2021-defending} proposed ADFAR, which trains an anomaly detector for identifying adversarial examples and performs frequency-aware randomization only for the adversarial inputs. The authors observe that this scheme preserves adversarial robustness without sacrificing the task performance. We adopt ADFAR, Sample Shielding, and SAFER as the other set of baselines for our work. 
\section{Learning to Remove Adversarial Perturbations}
\begin{figure}
\includegraphics[trim={0mm 0mm 0cm 0cm},clip, width=\linewidth]{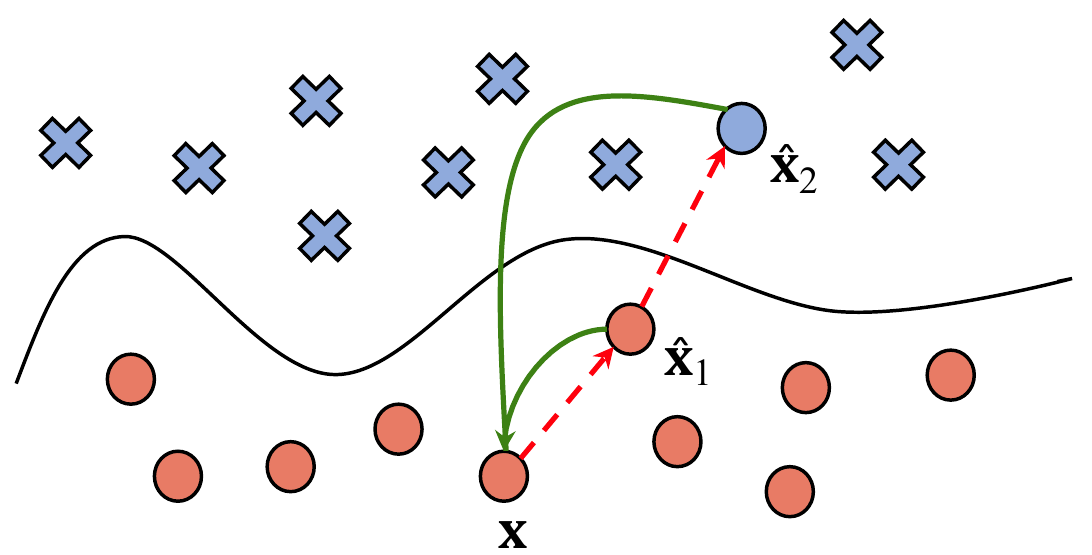}
\caption{Geometry of an adversarial attack. An attacker moves a point $\x$ (red circle) to the other side of the decision boundary along the red dotted arrows. $\hat{\x}_1$ represents the example after one change, $\hat{\x}_2$ denotes the final adversarial example. Our model~\modelname\ aims to restore these points to their original position (represented by the green curved arrows).
}
\label{fig:geometry}
\end{figure}
%
%
%
As mentioned in~\cref{sec:related_work}, most existing methods that operate on the textual input rely on randomization to remove any adversarial perturbations. We present a model that \textit{learns} to remove these perturbations by rewriting the text. 
Although similar to paraphrasing in terms of the task interface, our goal is different; we focus on removing the perturbation instead of just preserving the meaning. 
The rest of this section formalizes this intuition.

\subsection{Notation}


Given a sequence of tokens $\mathbf{x}$, suppose we have a trained classifier with parameters $\btheta$, which maps $\mathbf{x}$ to the output label $y$ from the label space $\mathcal{Y}$ as
$$
y = \argmax_{y_i \in \mathcal{Y}} P_{\btheta}(y_i | \mathbf{x})
$$
When the classifier makes a correct prediction, $y = y^{\ast}$, the true label for that input.
%
%
For a successful adversarial attack, the adversary takes the input sequence $\x$ and produces a perturbed variant $\mathbf{\hat{x}}$ by making a small change to $\x$ such that the prediction made by the model is incorrect:  
$$
 \argmax_{y_i \in \mathcal{Y}} P_{\btheta}(y_i | \mathbf{\hat{x}}) \ne y^{\ast}
$$
Additionally, the adversary ensures that the perturbation is \textit{meaningful} and \textit{imperceptible}.\footnote{Most attacks ensure this by enforcing constraints on the part-of-speech tags of the replaced words as well as maintaining the fluency through an LM perplexity score.} 

Typically, this perturbation is achieved via an iterative process, during which the adversary makes incrementally small modifications to the input~\citep{ebrahimi-etal-2018-hotflip,jin2020bert}. 
%
Assume that the adversary makes $k$ successive changes to the input
taking the original input $\x$ to the final adversarial variant $\mathbf{\hat{x}}_k$, represented as follows
$$
\x \rightarrow \mathbf{\hat{x}}_1 \rightarrow \mathbf{\hat{x}}_2 \rightarrow ... \rightarrow \mathbf{\hat{x}}_k
$$
To construct an adversarial input, the adversary selects perturbations such that the probability of the true label decreases after each successive modification until the required incorrect prediction is achieved. In other words, the adversary wants 
\begin{align}
    P_{\btheta}(y^{\ast} | \x) 
    > P_{\btheta}(y^{\ast}| \hat{\x}_1) 
    > \cdots 
    > P_{\btheta}(y^{\ast} | \hat{\x}_k) \label{eq:probs}
\end{align}
%
%
In summary, given an input sequence $\mathbf{x}$, an adversary keeps introducing small changes ($\mathbf{x} \rightarrow \mathbf{\hat{x}}_1$, etc.), each of which reduces the predicted probability of the correct label $y^{\ast}$.~\Cref{fig:geometry} shows a visual representation of the adversarial process, which moves the example marked $\x$  to the wrong side of the decision boundary by following the red arrows. 
%
\subsection{Training the Text Rewriter}
As we saw in~\cref{sec:related_work}, the most common strategy to counter such adversarial attacks is to retrain the model with new parameters $\btheta_{new}$ such that
$$
\argmax_{y_i \in \mathcal{Y}} P_{\btheta_{new}}(y_i | \mathbf{\hat{x}}_k) = y^{\ast}
$$
In this work, our objective is to keep the model parameters $\btheta$ unchanged and instead manipulate the input $\x$ by rewriting it. 
To this end, we define~\modelname\ that intercepts and rewrites potentially adversarial inputs. \modelname\ is a text-to-text transducer, which we will denote by $\mathcal{T}_{\bphi}$ with its own trainable parameters $\bphi$.

To effectively counteract adversarial inputs, the transformation function $\mathcal{T}_{\bphi}$, must transform the input $\x$ such that the task classifier makes the correct prediction on the transformed text:
$$
\argmax_{y_i \in \mathcal{Y}} P_{\btheta}(y_i | \mathcal{T}_{\bphi}(\mathbf{\hat{x}}_k)) = y^{\ast}
$$
We can guarantee this outcome by simply training $\mathcal{T}_{\bphi}$ to ensure that $\mathcal{T}_{\bphi}(\mathbf{\hat{x}}_k) = \x$. In other words, our goal is to learn a transformation function, $\mathcal{T}_{\bphi}$, that is capable of \textit{undoing} or \textit{reversing} the impact of adversarial modifications.
%
%
However, merely training the rewriter to reverse the final step $\mathbf{\hat{x}}_k$ may not be sufficient because $\mathbf{\hat{x}}_k$ is produced through a series of small changes. Therefore, in addition to undoing $\mathbf{\hat{x}}_k$, the rewriter should also be able to reverse each intermediate step. 
This strategy is based on the intuition  
that each successive change made to the input $\x$ in constructing an adversarial input  $\mathbf{\hat{x}}_k$ is in itself adversarial; 
 all intermediate changes 
 decrease the probability of the true label and are thus undesirable (\cref{eq:probs}). Green curved arrows in~\cref{fig:geometry} show the task of the rewriter.~\Cref{fig:adversarial_illustration} in the appendix shows an example of this process.

In summary, any adversarial modification made to the input at any stage should be reversed by~\modelname\ , i.e.,
\begin{equation}
    \mathcal{T}_{\bphi}(\mathbf{\hat{x}}_i) = \x, \forall i \in \{1, k\}
    \label{eq:adversarial}
\end{equation}
Finally, on non-adversarial inputs, we do not need to make any changes, and the function $\mathcal{T}_{\bphi}$ should therefore act as an identity function on these inputs:
\begin{equation}
    \mathcal{T}_{\bphi} (\x) = \x
    \label{eq:clean}
\end{equation}

\paragraph{Training Details} We use the T5 model~\citep{raffel2020exploring} as the starting point for our text rewriter~\modelname. Since we need adversarial examples to train our rewriter, we follow~\citet{bao-etal-2021-defending} and choose TextFooler~\citep{jin2020bert} for generating these examples on the whole training set. 
The training data for \modelname\ consists of input-output pairs of the form $(\mathbf{\hat{x}}_i, \mathbf{x})$: as described in~\cref{eq:adversarial}, for every adversarial modification $\mathbf{\hat{x}}_i$, including the ones with intermediate changes, and the original, unperturbed sequence  $\mathbf{x}$ is the desired output. In addition, as per~\cref{eq:clean}, the training data also includes unperturbed examples of the form $(\textbf{x},  \textbf{x})$.~\Cref{fig:rewriter_illustration} shows an illustrative example.

We train the \texttt{base} variant of the T5 model for 5 epochs with the starting learning rate of $5\times 10^{-5}$.
More details on the hyperparameters are provided in the appendix~\cref{app:hyper}. We use the Transformers ~\citep{wolf-etal-2020-transformers} for our implementation.

\section{Experimental Setup}
In this section, we will detail the datasets we use for our experiments, the baseline defense mechanisms, and the adversarial attacks they will be pitted against, and the three metrics we will use to compare the defense methods.

\subsection{Datasets}
We evaluate our proposed defense on four text classification datasets. 

\noindent \textit{\underline{Stanford Sentiment Treebank (SST-2)}} The SST-2 dataset is used for sentiment classification~\citep{socher-etal-2013-recursive} among two labels: \textit{positive}, and \textit{negative}. We use the splits from the GLUE benchmark~\citep{wang2019glue}; we use the validation set for reporting our results since the test set is not available publicly.

\noindent\textit{\underline{Rotten Tomatoes Movie Reviews}}~\citep[MR,][]{pang-lee-2005-seeing} Similar to SST-2 task, the goal is to predict a movie review's sentiment (\textit{positive} vs. \textit{negative}). We use the official test set for evaluation.

\noindent \textit{\underline{AG News}}~\citep{zhang2015character} This is a news classification dataset with four possible labels (\textit{sports}, \textit{world}, \textit{science/technology}, \textit{business}). The test set contains 7600 examples and since it can take a long time for robustness evaluation across all seven models and the five attackers,
we randomly choose 1000 examples for our evaluation set. 

\noindent\textit{\underline{Multi-Genre Natural Language Inference}}~\citep[MNLI,][]{williams-etal-2018-broad}
This is a standard dataset for Natural Language Inference (NLI) where the goal is to determine the inferential relation between a premise and a hypothesis. The dataset requires sentence-pair classification among three labels (\textit{entailment}, \textit{neutral}, and \textit{contradiction}). Again, we sample 1000 instances from the \texttt{validation-matched} subset for evaluation.
\begin{table}[]
\centering
\begin{tabular}{lccc}
\toprule
\textbf{Dataset}   & \textbf{\# Avg. words}      & \textbf{\# Labels}      & \textbf{Size} \\ \midrule
\textbf{SST-2}       & 9.4            & 2                    & 68K       \\
\textbf{MR}            & 21.6           & 2                    & 11K       \\
\textbf{AGNews}        & 44.1           & 4                    & 127K      \\
\textbf{MNLI}       &    33.9       & 3                    & 433K       \\
 \bottomrule
\end{tabular}
\caption{Summary of the datasets used in this paper. The three columns show the average number of words in the input, the number of labels, and the total size of the dataset. For MNLI, the input size is calculated by concatenating premise and hypothesis.
}
\label{tab:dataset_stats}
\end{table}
\subsection{Baselines and Adversarial attacks}
\paragraph{Baselines} We compare our model with a number of  baselines: Adversarial Training~\citep[AT,][]{alzantot-etal-2018-generating}, SHIELD~\citep{le-etal-2022-shield}, SAFER~\citep{ye-etal-2020-safer}, SampleShielder~\citep{rusert-srinivasan-2022-dont}, and ADFAR~\citep{bao-etal-2021-defending}. SAFER and SampleShielder are input randomization methods, while AT, and SHIELD require model retraining. ADFAR requires retraining the model and also uses input randomization. 
We could not compare the results with DISP~\citep{zhou-etal-2019-learning} as we were not able to run their implementation. We have provided more details in the appendix~\cref{sec:appendix}.

\paragraph{Adversarial Attacks}
We use the open source 
toolkit 
TextAttack~\cite{morris2020textattack, morris2020reevaluating} to evaluate all models on five black-box adversarial attacks. TextFooler~\citep{jin2020bert}, PWWS~\citep{ren-etal-2019-generating}, and BAE~\citep{garg2020bae} attack at the word-level, DeepWordBug~\citep[DWB, ][]{gao2018black} attacks at the character-level, and TextBugger~\citep{li2018textbugger} attacks at both word and character-level. TextFooler and PWWS use counter-fitted word embeddings~\citep{mrksic-etal-2016-counter}, while BAE uses the BERT as a masked language model~\citep{devlin-etal-2019-bert} to find the best word replacements. We provide an example of each of them in~\cref{tab:adversarial_attacks} in the appendix. 

\noindent We perform our main experiments with a BERT-base classifier as the victim
model with hyperparameters as suggested by~\citet{devlin-etal-2019-bert}.

\subsection{Evaluation}
\paragraph{Evaluation Metrics}
We measure the quality of the defense methods using three metrics, namely Clean Accuracy (Clean Acc.), Adversarial Accuracy (AA), and Average number of queries (\#Q). 

\noindent\textit{\underline{Clean Accuracy}} is the accuracy of the model on clean non-adversarial inputs, measured on the original validation or test sets. A model
that retains the clean accuracy of the original model is desirable.

\noindent\textit{\underline{Adversarial Accuracy (AA)}} The Attack Success Rate (ASR) of an attack is the percentage of instances where the attack algorithm successfully constructs an adversarial example. A defense method that makes a model more robust results in a lower ASR. We report the Adversarial Accuracy of the defense methods, defined as $100 - \text{ASR}.$

\noindent\textit{\underline{Average Number of Queries (\#Q)}} is the measure of the cost for an attacker, and is the average number of forward passes (queries) to the model by the attacker. On average, a more robust defense method requires more queries.

\paragraph{Evaluation Protocol}
The adversarial accuracy
depends on the number of queries an attacker is allowed to perform - a lower query budget entails a higher AA. There is currently no established protocol for evaluating the adversarial robustness of text classification systems. In this study, we do not impose a restriction on the number of queries allowed to the attacker, resulting in the most challenging conditions for the defense methods.


\section{Main Results}
\label{sec:results}
\begin{table*}[!ht]
\centering
\resizebox{\textwidth}{!}{
\begin{tabular}{@{}llrrrrrrrrrrr@{}}
\toprule
\multirow{2}{*}{\textbf{Dataset}}                   & \multirow{2}{*}{\textbf{Defense}} & \multirow{2}{*}{\begin{tabular}[c]{@{}l@{}}\textbf{Clean}\\ \textbf{Acc.}\end{tabular}} & \multicolumn{2}{c}{\textbf{TextFooler}} & \multicolumn{2}{c}{\textbf{TextBugger}} & \multicolumn{2}{c}{\textbf{BAE}} & \multicolumn{2}{c}{\textbf{PWWS}} & \multicolumn{2}{c}{\textbf{DWB}} \\
 \cmidrule(lr){4-5} \cmidrule(lr){6-7} \cmidrule(lr){8-9} \cmidrule(lr){10-11} \cmidrule(lr){12-13}
                                           &                          &                                                                       & AA            & \#Q            & AA            & \#Q            & AA         & \#Q        & AA         & \#Q         & AA         & \#Q        \\
\midrule 
{\multirow{7}{*}{SST-2}} & None                     & 92.4                                                                  & 4.8           & 95.4           & 31.3                & 49.3                &33.9            &60.4             &  13.4          & 143.1            & 18.6           & 34.7            \\
\cdashline{2-13}
\multicolumn{1}{c}{}                       & AT                  & 88.4                                                                  & 5.7           & 91.6           &  23.1             & 46.3                & 34.6           &  61.8          &13.2            &139.4             &10.1            &32.3            \\
\multicolumn{1}{c}{}                       & SHIELD                   & 88.8                                                                  & 6.6           & 90.9           & 25.1              & 51.4                & 28.5            & 61.3            & 13.6            & 137.1             &9.7            & 33.2            \\
\multicolumn{1}{c}{}                       & SAFER                    & 89.3                                                                  & 8.7           & 91.9           & 27.7              & 48.4                & 36.3            & 62.2            & 16.2            & 138.8             & 16.5            & 32.4            \\
\multicolumn{1}{c}{}                       & SampleShielder           & 76.8                                                                  & 6.6          & 97.1          & 25.7              & \textbf{58.4 }               & 28.8            & 66.2            & 17.7           & 143.8             &  17.5           & 36.2            \\
\multicolumn{1}{c}{}                       & ADFAR                    & 89.9                                                                  & 19.5          & 115.4          & 29.3              & 58.1                & \textbf{37.1}            & \textbf{68.7}           &  20.9           & 142.7            & 22.8            & 36.1            \\
\multicolumn{1}{c}{}                       & ~\modelname           & \textbf{92.0 }                                                                 & \textbf{24.0}          & \textbf{136.7}          & \textbf{40.5}              & 54.3                & 34.2            & 60.4            & \textbf{22.9}            & \textbf{150.1}             & \textbf{25.3}            & \textbf{38.0 }           \\
\midrule
\multirow{7}{*}{MR}                        & None                     & 84.2                                                                  & 10.7          & 117.7           & 37.3              & 56.1                & 38.4            & 64.4            &  18.7           & 150.0            & 22.3           & 40.5            \\
\cdashline{2-13}
                                           & AT                  & 84.2                                                                  & 11.3          & 118.6          &  34.3             & 54.8                & 35.9           & 65.8            &19.2            &151.1             & 18.1           & 38.2            \\
                                           & SHIELD                   & 82.1                                                                  & 12.1          & 98.7           & 22.3              & 60.8                & 27.4            & 65.6            & 18.2           & 141.7             & 18.7            & 37.2            \\
                                           & SAFER                    & 85.5                                                                  & 3.7           & 88.1           & 23.4              & 49.3                & 33.4            & 59.8            & 10.6            & 142.0             & 16.0            & 34.0           \\
                                           & SampleShielder           & 76.2                                                                  & 12.1          & 105.5          &26.5               &58.2                & 27.3           &  61.7           & 21.4            & 150.7             & 24.3          & 39.7            \\
                                           & ADFAR                    & 82.4                                                                  & 17.5          & 120.5          & 26.0              & 59.6                & 31.4           & 65.5            & 23.0           & 148.8             & 22.6            & 38.2            \\
                                           &~\modelname              & \textbf{84.3}                                                                  & \textbf{21.1}          & \textbf{140.2}          & \textbf{45.7}               & \textbf{61.0}                & \textbf{38.6}            & \textbf{65.8}            & \textbf{26.4}            & \textbf{154.2}             & \textbf{32.5}            & \textbf{43.6}            \\
\midrule
\multirow{7}{*}{MNLI}                      & None                     & 83.5                                                                  & 1.1           &  81.3              &   4.2            &   54.1             &  19.3          &  59.2          &   2.4         &  188.3           &  4.2          & 41.5           \\
\cdashline{2-13}
                                           & AT                  & 80.8                                                                  & 2.7           &  105.0              &   6.3            &  59.3              & 20.7           & 62.5           &3.5            &190.4             &6.9            &41.7            \\
                                           & SHIELD                   &  79.5                                                                     & 2.9               & 103.8                & 6.7               & 60.2                & 20.9            & 63.1            & 3.4            & 191.1             & 7.9            & 44.0            \\
                                           & SAFER                    &   78.0                                                                     & 1.7               & 101.3                & 10.3               & 58.8                &  24.5           & 62.9            & 5.3            & 196.7             & 8.3            & 44.1            \\
                                           & SampleShielder           & 41.4                                                                       & \textbf{17.5}               & \textbf{178.3}                & \textbf{17.2}               & \textbf{102.2}                & \textbf{41.7}            & \textbf{100.1}            & \textbf{26.1}            & \textbf{231.2}             & \textbf{19.9}             & \textbf{57.0}            \\
                                           & ADFAR                    & 78.1                                                                  & 10.5          & 117.8               & 7.6               & 64.3                & 16.3           & 61.6          & 11.0            & 200.7             &9.4            &44.9            \\
                                           &~\modelname             & \textbf{83.0}                                                                  & 16.1          & 158.2              & 9.7              & 67.3                & 20.4            & 61.5            & 10.9            &  195.2            & 9.5            & 45.4            \\
\midrule
\multirow{7}{*}{AGNews}                    & None                     & 94.9                                                                  & 18.2          & 334.1          & 47.7               & 180.9                &  84.8           & 116.8            & 43.2            & 353.0             & 38.9            & 110.1            \\
\cdashline{2-13}
                                           & AT                  & 94.1                                                                     & 19.1             & 379.2              & 49.1               & 189.7                & 83.4            & 117.1            & 44.1            & 355.6             & 39.7            & 114.2            \\
                                           & SHIELD                   & 92.4                                                                     & 20.1             & 385.3              & 51.7               & 190.9                & 81.8            & 114.4            & 44.9            & 359.4             & 39.7            & 112.4            \\
                                           & SAFER                    & 91.2                                                                  & 15.7          & 280.6          &  33.6             & 156.6                & 78.8            & 119.9            & 45.8             & 361.2             & 40.8            & 114.7            \\
                                           & SampleShielder           & 90.8                                                                  & 52.6          & 425.6          &     56.7          & 216.9                &   84.4          & 119.5            & 49.8            & 365.4             & 41.6            & 115.4            \\
                                           & ADFAR                    & 92.4                                                                  & 58.3          & 422.2          & 52.5               & \textbf{245.1}                & 79.7            & \textbf{136.3}            & 45.9            & 368.4             & 47.1            & 115.8            \\
                                           &~\modelname              & \textbf{94.7}                                                                  & \textbf{73.0}          & \textbf{520.0}          & \textbf{63.9}               & 222.9                & \textbf{87.3}            & 123.5            & \textbf{63.9}            & \textbf{375.2}             & \textbf{49.7}            & \textbf{117.3}           \\
\bottomrule
\end{tabular}
}
\caption{Results comparing model robustness using the clean accuracy (\%) and adversarial accuracy (\%) on the five adversarial attacks: 
None indicates the BERT model without any defense and therefore acts as a baseline model. Notably, our  model~\modelname\ yields superior results across the board without significant drop in clean accuracy. 
}
\label{tab:resultsAll}
\end{table*}
~\Cref{tab:resultsAll} shows the results for the defense methods on all four datasets. 
Additionally,~\cref{tab:resultsSummary} in the appendix summarizes the results in terms of average improvements for the five adversarial attacks.

As observed from the table, our proposed method~\modelname\  provides a consistent and substantial improvement in terms of adversarial robustness over the baselines. We find that there is a trade-off between clean accuracy and adversarial robustness for all the models, aligning with the findings of~\citet{raghunathan2020understanding}. 
The results show that~\modelname\ maintains the highest level of clean accuracy on all datasets except MR, where SAFER improves it by more than 1\%, but does so at the cost of making the model less robust.

The most formidable baseline is ADFAR, which employs an anomaly detector to identify adversarial inputs and uses input randomization for handling adversarial instances. Our method substantially outperforms ADFAR on all settings except one. Furthermore, we observe that SampleShielder performs well on AGNews but not on other datasets. This can be attributed to the fact that SampleShielder randomly removes parts of the input before making a prediction. This is effective for tasks with longer inputs and simpler semantics (such as topic classification on AGNews), but does not work for others where removing parts of the input can alter the label. Additionally, while SampleShielder provides the best adversarial accuracies on the MNLI dataset, the clean accuracy is almost close to random. Our proposed model~\modelname\, on the other hand, provides the best balance between adversarial and clean accuracies.

\subsection{Results against other attack types}
Several defense methods, including ours, utilize adversarial examples from one attack type to train their models. The true effectiveness of adversarial defenses is determined when they are tested against previously unseen adversarial attacks. Our evaluation using four other attacks, excluding TextFooler, accomplishes this. Each of these attacks differ from TextFooler in one or more aspects. For example, while TextFooler is a token-level attack, DeepWordBug (DWB) is a character-level attack. TextBugger, on the other hand, is a multi-level attack, capable of attacking at both token and character level. BAE replaces words uses a BERT MLM while TextFooler uses GloVe word embeddings. PWWS, in comparison, employs a different algorithm for token replacement.

From ~\cref{tab:resultsAll}, we observe that, as compared to the baselines,~\modelname\  provides significant improvements in robustness against other attacks. Notably, while~\modelname\ is only trained against synonym substitutions from TextFooler, it is able to generalize to other attacks that operate at the character level. Lastly, the improvement against BAE is less than that against other attacks. We hypothesize that this is due to the fact that BAE employs a BERT language model for word replacements, which is different from the technique used by TextFooler. 

\subsection{Transferability to other classifiers}
\begin{table}[]
\begin{tabular}{lrr}
\toprule
             & Clean Acc & Adv. Acc. \\ \midrule
BERT         & 92.4      & 4.8       \\
+ ~\modelname & 92.0      & 24.0      \\
RoBERTa      & 94.1      & 5.0       \\
+ ~\modelname & 93.7      & 25.1      \\
DistilBERT   & 90.0      & 2.9       \\
+ ~\modelname & 89.5      & 17.8      \\
ALBERT       & 91.1      & 4.2       \\
+ ~\modelname & 90.4      & 19.0      \\ \bottomrule
\end{tabular}
\caption{Transferability to other models.The ~\modelname\ is trained for defending the BERT model and is evaluated for its ability to defend other models without retraining. The results are shown for the SST-2 dataset for its validation subset. }
\label{tab:transferability_models}
\end{table}

As mentioned previously, one motivation for using a separate robustness module like ours is that it can be transferred to other text classification models without retraining the rewriter. 
We use~\modelname\, which was trained to remove adversarial perturbations for the BERT classifier on the SST-2 dataset and employ it, without retraining, to remove adversarial perturbations for other classifiers on the same dataset. We assess the transferability of~\modelname\ against three classifiers, namely: RoBERTa~\citep{liu2019roberta}, DistilBERT~\citep{sanh2019distilbert}, and ALBERT~\citep{lan2019albert}. 

The results for evaluation against TextFooler are presented in~\cref{tab:transferability_models}. We observe that~\modelname\ is effective in enhancing adversarial robustness for models other than BERT. Importantly, this improvement is achieved without much drop in performance on the clean examples ($< 1\%$ in all cases). On average,~\modelname\ improves adversarial accuracy by 16.6\% across the three models. Surprisingly, the improvement for the RoBERTa model is even more pronounced than that for the BERT model. We hypothesize that this transferability from~\modelname\ is due to two factors. First, adversarial attacks often result in similar adversarial changes, particularly for the same dataset. Second, previous research has demonstrated that adversarial examples transfer across classifiers for the same task~\citep{papernot2016transferability,DBLP:conf/iclr/LiuCLS17}.


\subsection{Transferability to other tasks and datasets}
\begin{table}[]
\begin{tabular}{lrr}
\toprule
             & Clean Acc & Adv. Acc. \\ \midrule
BERT-SST2        & 92.4      & 4.8       \\
+ ~\modelname & 92.0      & 24.0      \\
\hdashline
MR      & 84.2      & 10.7       \\
+ ~\modelname & 84.2      &29.3     \\
AGNews   & 94.2      & 18.2       \\
+ ~\modelname & 93.1      & 30.8      \\
MNLI & 83.5 & 1.1 \\
+ ~\modelname & 83.2 & 2.8\\
\bottomrule
\end{tabular}
\caption{Results comparing transferability of ~\modelname\ to other tasks. The~\modelname\ trained for the BERT model on SST-2 dataset is evaluated for BERT classifiers on other datasets without retraining.}
\label{tab:transferability_datasets}
\end{table}

As explained in the previous section,~\modelname\ allows for its application to tasks and datasets for which it was not trained. We now assess the transferability of our method with respect to other tasks and datasets. We use the~\modelname\ trained for sentiment classification on SST-2 using BERT and apply for the BERT model trained on other datasets. We perform this evaluation on three datasets, namely MR, AGNews, and MNLI. 

We present the results in the~\cref{tab:transferability_datasets}. We find that our model~\modelname\ exhibits strong transferablity for other datasets. Again, as with previous results, we see only small drops in performance on non-adversarial inputs. The favorable results on the MR dataset shows that~\modelname\ effectively transfers for a different dataset of the same task. Note that the improvement in adversarial accuracy for MR is even higher than a model that is specifically trained for removing adversarial perturbations for the MR dataset (see~\cref{tab:resultsAll}). This is explained by the fact that the MR dataset is much smaller and thus the~\modelname\ trained on that dataset has fewer adversarial instances to learn from (10k vs. 67k). We notice more than 12\% increase in adversarial accuracy on the AGNews dataset. This is perhaps most surprising, since not only the task is semantically different with different set of classes, but the domain of the dataset is also different (movies vs. news). On the MNLI dataset though, we notice only small improvement, perhaps because it is a semantically harder task. 

In summary, our proposed model~\modelname\ transfers across both models and datasets. This observation can motivate the training of a single rewriter module for all tasks and datasets. The benefits of such an approach are two-fold. First, since the defense capability transfers across models, a single shared model could be more robust than the individual ones. Second, having a single shared is more practical as it reduces the overhead in deployment of~\modelname. We leave the exploration of this shared rewriter approach to future work.

\subsection{Effect of the model size}
\begin{table}[]
\centering
\resizebox{\columnwidth}{!}{
\begin{tabular}{lccc}
\toprule
 Model              & Params. & Clean Acc & Avg. AA \\ \midrule
\texttt{t5-small}   &  60M    & 92.4      &  21.9      \\
\texttt{t5-base}    &  220M   & 92.0      &  29.4     \\
\texttt{t5-large}   &  770M   & 92.4      &  37.8      \\
\texttt{t5-3b}        &  3B      &  92.1   &  45.9\\ 
 \bottomrule
\end{tabular}
}
\caption{Results comparing the robustness w.r.t the size of ~\modelname\  model. All the models are evaluated on SST-2 with the BERT-\texttt{base} classifier. We report here the averaged adversarial accuracies for compactness. Please refer to the appendix~\cref{tab:model_size_detailed} for detailed numbers.}
\label{tab:model_size}
\end{table}
For all experiments in previous sections, we used the \texttt{base} variant of the T5 model for training ~\modelname. We now investigate the effect of the size of the rewriter module on the adversarial robustness. For the SST-2 dataset, we train four variants of ~\modelname\  with different sizes: \texttt{t5-small}, \texttt{t5-base}, \texttt{t5-large}, \texttt{t5-3b}. The results are shown in~\cref{tab:model_size}. We observe that with increased size, the rewriter module defends the classification model more robustly. 

\subsection{Pre-training the Rewriter}
One additional benefit of having a separate rewriter module is that we can pre-train the rewriter without using any task-specific datasets. We demonstrate this approach by artificially constructing a training corpus using the Wikipedia text. Specifically, we sample 100k sentences from the English Wikipedia and randomly substitute 15\% of the words in each of those sentences with one of the neighbors from the GloVe embedding space~\citep{pennington-etal-2014-glove}. The pre-training task for the rewriter is to simply \textit{reverse} this perturbation by generating the original unperturbed sentence. Note that this setup is close to but does not perfectly simulate the actual adversarial attack scenario, 
as the perturbations used in the latter are chosen with greater precision. We observe that this pre-training improves the ~\modelname\ by more than 2.5\% in terms of adversarial accuracy without any significant decrease in clean accuracy. Due to space constraints, the results are shown in~\cref{tab:wiki_pretraining} in the appendix .  

\subsection{Latency at Inference}
One limitation of our proposed strategy is that it utilizes two neural models to make predictions, hurting the overall inference time. We measure latency for each of the models by averaging their inference time over 200 examples (100 clean + 100 adversarial).
We observe that~\modelname\ is slower than model retraining approaches (22.0 ms for SHIELD vs. 95 ms
for~\modelname), while being faster or competitive with input randomization methods. SAFER is the slowest of all since it performs averaging over a large number of candidate synonyms. 

One possible approach to reduce inference time could be to use more efficient text generation models like non-autoregressive text generation~\citep{gu2018non}. Moreover, a method based on text-editing can also be promising~\citep{malmi-etal-2022-text}. We leave these explorations to the future work.












\section{Conclusion}
In this paper, we explore a novel strategy to defend against textual adversarial attacks that does not require model retraining. 
Our proposed model, \hbox{\modelname} intercepts and rewrites adversarial inputs to make them non-adversarial for a downstream text classifier. 
We perform experiments on four text classification datasets and test its effectiveness against five adversarial attacks. The results suggest that, in comparison with baselines, our proposed approach is not only more effective against adversarial attacks but is also better at preserving the task accuracies. Moreover, when used as a pluggable module,~\modelname\ shows great transferability to new models and datasets---on three new datasets, it improves adversarial accuracy by 10.9\% on average. We expect the future work to focus on improving inference time latency by using more sophisticated text generation methods.

\section{Limitations}
This work is subject to two limitations. First, our experiments were restricted to text classification tasks and we did not evaluate if our methods can effectively defend against adversarial attacks for other tasks like QA, etc.~\citep{jia-liang-2017-adversarial}. It therefore remains unexplored if our conclusions transfer beyond the text classification tasks. 

Second, the primary contribution of our work,~\modelname\  relies on using a language model like T5, which is trained on large amount of text in English. It is possible that our approach is not as effective for languages where such a model is not freely available. Additionally, in this work, we did not explore the impact of large language model pretraining on our results.

\section{Ethical Considerations}
This work is concerned with protecting or defending against adversarial attacks on text classification systems. For modeling, our method~\modelname\ uses another neural network based language model T5~\citep{raffel2020exploring}. This means the~\modelname\ can itself be attacked by an adversary. 
We believe that attacking a pipelined model such as~\modelname\ is not straightforward for the following two reasons. First, performing an adversarial attack on a model typically requires access to output scores from that model. Since ATINTER is used in a pipeline with a task classifier, the attacker can never get access to ATINTER’s output scores. This adds an additional layer of complexity for the adversary. Second, targeted adversarial attacks on sequence-to-sequence models (such as ATINTER) are much less prominent and it is generally more difficult to make small alterations in the input without forcing a more significant change in the textual output~\citep{cheng2020seq2sick,tan-etal-2020-morphin}.
Nevertheless, we have not explored this possibility and therefore recommend practitioners interested in using this work to carefully check for this. 

Additionally, the experiments were only performed on four text classification datasets. Although we expect our method to be effective for other classification tasks like Toxicity detection, Hate Speech identification, but considering the sensitive nature of these applications, we urge the practitioners to first comprehensively evaluate our work on those tasks before deploying in a real world scenario.

For all our experiments, we used pre-established and published datasets, which do not pose any serious ethical concerns. For transparency and reproduciblity, we will make our code publicly available. 

\section*{Acknowledgements}
The authors thank Bloomberg’s AI Engineering team, especially Umut Topkara and Anju Kambadur for helpful feedback and directions. We would also like to thank members of the Utah NLP group for their valuable insights, and the reviewers for their helpful feedback. This work was supported in part by the National Science Foundation under Grants \#1801446, \#1822877, \#2007398 and \#2217154.  Any opinions, findings, and conclusions or recommendations expressed in this material are those of the authors and do not necessarily reflect the views of the National Science Foundation.

\bibliography{anthology,custom}
\bibliographystyle{acl_natbib}

\appendix

\section{Appendix}
\label{sec:appendix}

\begin{table*}[]
\centering
\begin{tabular}{llll}
\toprule
\textbf{Attack} & \textbf{Type} & \textbf{Can humans} & \textbf{Example}  \\
  &  & \textbf{identify it?} &  
 \\ \midrule
TextFooler & word-level & NO & \underline{Org:} The child is at the beach. \\
           &            &    & \underline{Adv:}  The \textbf{youngster} is at the \textbf{shore}. \\
           
TextBugger & char-level, & YES & \underline{Org:} I love these awful 80's summer camp movies. \\
           & word-level &    & \underline{Adv:} I love these \textbf{aw ful} 80's summer camp movies. \\
           
BAE        & word-level & NO & \underline{Org:} The government made a quick decision. \\
           &            &    & \underline{Adv:} The \textbf{doctor} made a quick decision. \\
           
PWWS       & word-level & NO & \underline{Org:} E-mail scam targets police chief. \\
           &            &    & \underline{Adv:} E-mail scam targets police \textbf{headman}. \\

DeepWord   & char-level & YES & \underline{Org:} Subject: breaking news. would you ref inance ...\\
           &            &    & \underline{Adv:} su\textbf{jb}ect woul\textbf{g} y\textbf{uo} h\textbf{va}e an [OOV] ...  \\
\bottomrule
\end{tabular}
\caption{\textbf{Summary of the black-box adversarial attacks}: Comparing the adversarial attacks we use in this work along with related information such as attach type, human perceptibility, and an example input for each attack. The third column indicates whether a human can easily identify if textual input was modified or not based on grammar syntax, semantics, and other language rules.}
\label{tab:adversarial_attacks}
\end{table*}

\begin{table*}[]
\begin{tabular}{@{}lcccccccc@{}}
\toprule
Model                              & \multicolumn{1}{c}{\# Params} & \multicolumn{1}{c}{\begin{tabular}[c]{@{}c@{}}Clean \\ Acc.\end{tabular}} & \multicolumn{1}{c}{\begin{tabular}[c]{@{}c@{}}TextFooler\\ AA\end{tabular}} & \multicolumn{1}{c}{\begin{tabular}[c]{@{}c@{}}TextBugger\\ AA\end{tabular}} & \multicolumn{1}{c}{\begin{tabular}[c]{@{}c@{}}BAE\\ AA\end{tabular}} & \multicolumn{1}{c}{\begin{tabular}[c]{@{}c@{}}PWWS\\ AA\end{tabular}} & \multicolumn{1}{c}{\begin{tabular}[c]{@{}c@{}}DWB\\ AA\end{tabular}} & \multicolumn{1}{c}{Avg. AA} \\ \midrule
\texttt{t5-small} & 60M                           & \textbf{92.43}                                                                     & 11.29                                                                       & 31.39                                                                       & 33.62                                                                & 15.51                                                                 & 17.49                                                                &  21.86                           \\
\texttt{t5-base}  & 220M                          & 91.97                                                                     & 23.96                                                                       & 40.52                                                                       & 34.16                                                                & 23.06                                                                 & 25.31                                                                &   29.40                          \\
\texttt{t5-large}  & 770M                          & \textbf{92.43}                                                                     & 31.14                                                                       & 50.99                                                                       & 36.85                                                                & 30.4                                                                  & 39.45                                                                &       37.77                      \\
\texttt{t5-3b}     & 3B                            & 92.09                                                                     & \textbf{39.33}                                                                       & \textbf{57.53}                                                                            & \textbf{42.67}                                                                     & \textbf{35.1}                                                                  & \textbf{54.79}                                                                &  \textbf{45.88}                           \\ \bottomrule
\end{tabular}
\caption{Detailed Results with different sizes of the ~\modelname. AA stands for Adversarial Accuracy. The results shown here are for the SST-2 dataset and the BERT classifier.}
\label{tab:model_size_detailed}
\end{table*}

\subsection{Implementation Details}

\paragraph{SHIELD} We find that SHIELD is very sensitive to the $\tau$ hyperparameter involved. There is a strong trade-off between the clean accuracy and adversarial robustness for the change in $\tau$. For reporting the results, we try four values of $\tau = [1.0, 0.1, 0.01, 0.001]$ and report the results for the model that retains the accuracy the most. 

\paragraph{SAFER} Since SAFER is an input randomization method, the default implementation provides different results for different runs, although we do not see any substantial change in numbers. For reporting clean accuracy, we average it with the numbers obtained with each of the five attacks. Additionally, since SAFER aggregates predictions by considering a large number of candidates for random synonym replacements for each word it decides to perturb, we found it is not practical to run with number of candidates equal to 100 (used in the original implementation). Therefore, we report the numbers with $n=30$ in this paper.

\paragraph{ADFAR} The official ADFAR implementation~\footnote{\url{https://github.com/LilyNLP/ADFAR}} only provides instructions to reproduce results for MR dataset (more specifically for tasks with single input classification and with only two possible labels). We, therefore, modify the codebase to make it work for AGNews --that has four classes, and MNLI, where the task is sentence-pair classification. We will release our modified codebase for ADFAR for the community to reproduce these results.

\paragraph{DISP} We were not able to run the open-sourced implementation of DISP during our experiments~\footnote{\url{https://github.com/joey1993/bert-defender}}. We experimented with several different versions of both PyTorch and the transformers libraries but were still unsuccessful. More details can be found at a github issue we created at:~\footnote{\url{https://github.com/joey1993/bert-defender/issues/2}}.

\subsection{Hyperparameters for training~\modelname\ }
\label{app:hyper}
We list here the hyperparameters we used for training our model. 
\begin{enumerate}
    \item Learning Rate: We found that the learning rate of 5e-5 works best. We performed the learning rate search over the set [1e-5, 5e-5, 1e-6, 5e-6]. Also, we find the best learning rate for the SST-2 dataset and use the same for other datasets.
    \item Batch Size: For all our models, except \texttt{t5-3b}, we use the batch size of 16 during training. Wherever, the batch of 16 did not fit the GPU (for example on the 16GB V100), we use gradient accumulation to have the effective batch size of 16. We did not perform any hyperparameter search for batch size due to computational reasons.
    \item Sequence Length: Since examples in the SST-2 and MR datasets are smaller, we used the source and target side sequence lengths to both be 128. For AGNews, we use the sequence length of 512 and for MNLI, we use 256.
    \item Number of epochs: For all our models (except that involve wiki pre-training), we used 5 epochs. For the pre-training setup, we used 10 epochs. 
\end{enumerate}
For training \texttt{t5-3b}, we needed to use \texttt{DeepSpeed}~\footnote{\url{https://github.com/microsoft/DeepSpeed}} for our experiments. 

\subsection{Reproducibility Details}
\paragraph{Dataset Splits} We use the dataset splits from the Huggingface datasets repository.~\footnote{\url{https://huggingface.co/datasets}}. For datasets where we use a subsample of the test set, we use the random seed 1 to first shuffle and then evaluate on first 1000 instances.

\paragraph{Hardware} We run most of our experiments using the Nvidia V100 (32 GB) GPU. Some of the later experiments with T5-3b required even larger GPU RAM and therefore, I was able to use  Tesla A100 (40 GB VRAM) for last few experiments. Additionally, the servers had CPU: AMD EPYC 7513 32-Core Processor with CPU RAM 512 GB.
\begin{figure*}
    \centering
    \includegraphics[width=0.9\textwidth]{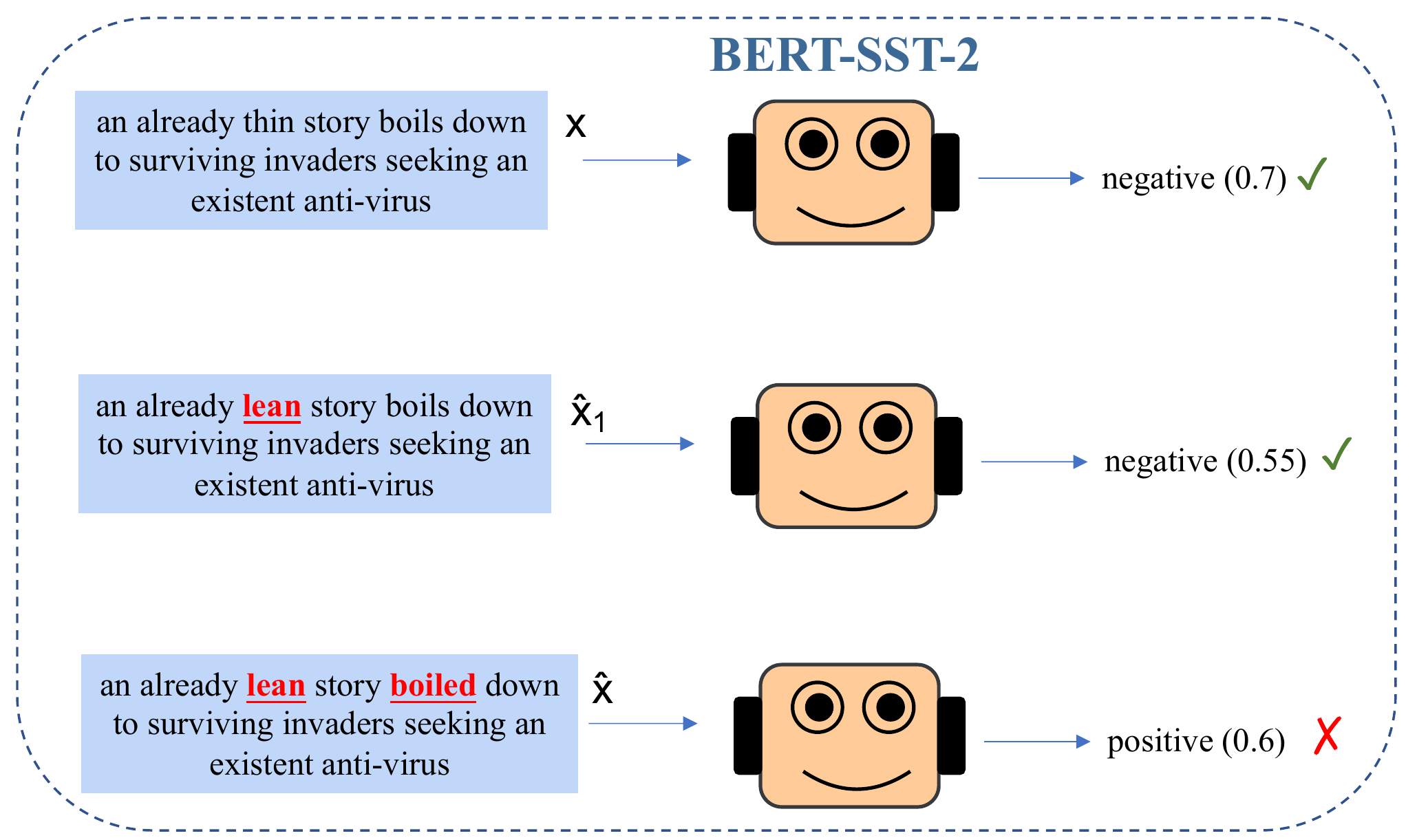} 
        \caption{Illustration of how an adversarial example is generated. Given a non-adversarial, correctly predicted sentence $\mathbf{x}$, a typical adversary makes a word substitution to first obtain $\hat{\x}_1$ which lowers the predictive probability and then makes another substitution to generate $\hat{\x}$, such that the predicted label becomes incorrect. }
        \label{fig:adversarial_illustration}
\end{figure*}
\begin{figure*}
    \centering
    \includegraphics[width=0.9\textwidth]{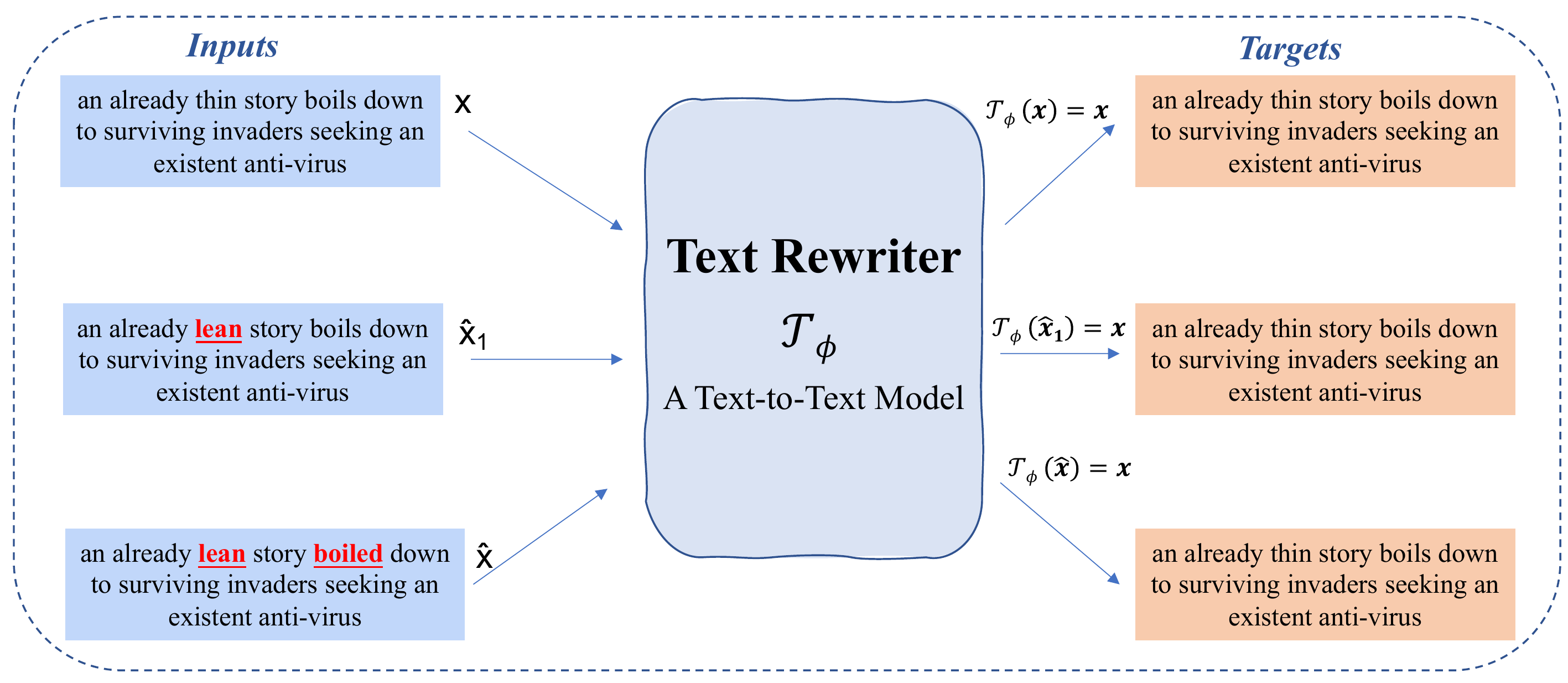} 
        \caption{Training Setup for the Text Rewriter. 
        For a non-adversarial example $\x$, the rewriter should learn to simply reproduce the same text as output. For the other two inputs ($\hat{\x}_1$, and $\hat{\x}$), the rewriter learns to restore the original input $\x$. Please refer to the main text for more details.
        }
        \label{fig:rewriter_illustration}
\end{figure*}

\begin{table}[]
\centering
\resizebox{\columnwidth}{!}{
\begin{tabular}{lccc}
\toprule
 Model               & Clean Acc & Adv. Acc. \\ \midrule
ADFAR~\citep{bao-etal-2021-defending}   & 89.9       & 19.5       \\
\modelname\ (pre-training only) & \textbf{92.3}& 9.6\\
\modelname\ (SST-2)       & 92.0      &  24.0     \\
 \ \ \ + pre-training   & 91.9      &  \textbf{26.5}      \\
 \bottomrule
\end{tabular}
}
\caption{Effect of Pretraining ~\modelname\ using wikipedia sentences. Results shown for the SST-2 dataset.}
\label{tab:wiki_pretraining}
\end{table}
\begin{table*}[!ht]
\centering
\begin{tabular}{lcccccccc}
\toprule
\textbf{Defense} & \multicolumn{2}{c}{\textbf{SST-2}}                                                                        & \multicolumn{2}{c}{\textbf{MR}}                                                                           & \multicolumn{2}{c}{\textbf{MNLI}} & \multicolumn{2}{c}{\textbf{AGNews}}                                                                       \\ \midrule
                 & \begin{tabular}[c]{@{}r@{}}Clean \\ Acc.\end{tabular} & \begin{tabular}[c]{@{}r@{}}Adv.\\ Acc.\end{tabular} & \begin{tabular}[c]{@{}r@{}}Clean \\ Acc.\end{tabular} & \begin{tabular}[c]{@{}r@{}}Adv.\\ Acc.\end{tabular} & \begin{tabular}[c]{@{}r@{}}Clean \\ Acc.\end{tabular} & \begin{tabular}[c]{@{}r@{}}Adv.\\ Acc.\end{tabular} &  \begin{tabular}[c]{@{}r@{}}Clean \\ Acc.\end{tabular} & \begin{tabular}[c]{@{}r@{}}Adv.\\ Acc.\end{tabular}\\
 \cmidrule(lr){2-3}  \cmidrule(lr){4-5}  \cmidrule(lr){6-7}  \cmidrule(lr){8-9}
AT          & -4.0 &  -4.0 & 0.0  & -1.5 & -2.7    & 1.8    & -0.7 & -1.0\\
SHIELD           & -3.6 & -3.7 & -2.1  & -5.1 & -4.0    & 1.9    & -2.5  & 1.2\\
SAFER            & -3.1 &  0.3& \textbf{1.3}  & -6.5  & -5.5    & 3.8    & -3.7 & -4.2  \\
SampleShielder   & -15.6 & -1.4 & -8.0 & -3.6 & -42.1 & -  & -4.1 & 9.4 \\
ADFAR            & -2.5 & 4.2 & -1.8 & -1.4 & -5.4    & 4.7    & -2.7 & 10.2 \\
~\modelname     & \textbf{-0.4} & \textbf{7.4} & 0.1 & \textbf{6.2} & -0.5 & \textbf{6.3}&  -0.2& \textbf{21.9}\\ \bottomrule
\end{tabular}
\caption{\textbf{Summary of the main results.} Absolute percentage change in Clean Accuracy and Adversarial Accuracies averaged over the five adversarial attacks.}
\label{tab:resultsSummary}
\end{table*}



\end{document}